%% file: main.tex
\documentclass{article} 
\usepackage{iclr2023_conference}
\usepackage{times}

\input{math_commands.tex}

\usepackage{hyperref}
\usepackage{url}

\author{%
 Paul Hilt \\
 Cell Biology and Biophysics Unit \\
 European Molecular Biology Laboratory \\
 \And
 Maedeh Zarvandi \\
 Cell Biology and Biophysics Unit \\
 European Molecular Biology Laboratory \\
 \And
 Edgar Kaziakhmedov \\
 Cell Biology and Biophysics Unit \\
 European Molecular Biology Laboratory \\
 Skolkovo Institute of Science and Technology \\
 \And
 Sourabh Bhide \\
 Director's Research \\
 European Molecular Biology Laboratory \\
 \And
 Maria Leptin \\
 Director's Research \\
 European Molecular Biology Laboratory \\
 European Molecular Biology Organization \\
 \And
 Constantin Pape \\
 Cell Biology and Biophysics Unit \\
 European Molecular Biology Laboratory \\
 \texttt{constantin.pape@informatik.uni-goettingen.de} \\
 \And
 Anna Kreshuk \\
 Cell Biology and Biophysics Unit \\
 European Molecular Biology Laboratory \\
 \texttt{anna.kreshuk@embl.de} \\
}

\usepackage[utf8]{inputenc} 
\usepackage[T1]{fontenc}    
\usepackage{hyperref}       
\usepackage{url}            
\usepackage{booktabs}       
\usepackage{amsfonts}       
\usepackage{nicefrac}       
\usepackage{microtype}      
\usepackage{xcolor}         
\usepackage{graphicx}
\usepackage{subfig}
\usepackage{amsmath}
\usepackage{wrapfig}
\usepackage[linesnumbered,ruled,vlined]{algorithm2e}

\title{Reinforcement learning for instance segmentation with high-level priors}

\iclrfinalcopy 
\begin{document}

\maketitle

\begin{abstract}
Instance segmentation is a fundamental computer vision problem which remains challenging despite impressive recent advances due to deep learning-based methods. Given sufficient training data, fully supervised methods can yield excellent performance, but annotation of groundtruth remains a major bottleneck, especially for biomedical applications where it has to be performed by domain experts. The amount of labels required can be drastically reduced by using rules derived from prior knowledge to guide the segmentation. However, these rules are in general not differentiable and thus cannot be used with existing methods. Here, we revoke this requirement by using stateless actor critic reinforcement learning, which enables non-differentiable rewards. We formulate the instance segmentation problem as graph partitioning and the actor critic predicts the edge weights driven by the rewards, which are based on the conformity of segmented instances to high-level priors on object shape, position or size. The experiments on toy and real data demonstrate that a good set of priors is sufficient to reach excellent performance without any direct object-level supervision. 
\end{abstract}

\section{Introduction}

Instance segmentation is the task of segmenting all objects in an image and assigning each of them a different id. It is the necessary first step to analyze individual objects in a scene and is thus of paramount importance in many computer vision applications. Over the recent years, fully supervised instance segmentation methods have made tremendous progress both in natural image applications and in scientific imaging, achieving excellent segmentations for very difficult tasks \cite{lee2017superhuman, chen2021scaling}. 

A large corpus of training images is hard to avoid when the segmentation method needs to take into account the full variability of the natural world. However, in many practical segmentation tasks the appearance of the objects can be expected to conform to certain rules that are known \emph{a priori}. Examples include surveillance, industrial quality control and especially medical and biological imaging applications where full exploitation of such prior knowledge is particularly important as the training data is sparse and difficult to acquire: pixelwise annotation of the necessary instance-level groundtruth for a microscopy experiment can take weeks or even months of expert time. The use of shape priors has a strong history in this domain \cite{osher2007geometric,delgado2014snakes}, but the most powerful learned shape models still require groundtruth \cite{oktay2018anatomically} and generic shapes are hard to combine with the CNN losses and other, non-shape, priors. 
For many high-level priors it has already been demonstrated that integration of the prior directly into the CNN loss can lead to superior segmentations while significantly reducing the necessary amounts of training data \cite{KERVADEC201988}. However, the requirement of formulating the prior as a differentiable function poses a severe limitation on the kinds of high-level knowledge that can be exploited with such an approach. Our contribution addresses this limitation and establishes a framework in which a rich set of non-differentiable rules and expectations can be used to steer the network training. 

To circumvent the requirement of a differentiable loss function, we turn to the reinforcement learning paradigm, where the rewards can be computed from a non-differentiable cost function. We base our framework on a stateless actor-critic setup \cite{DBLP:journals/corr/PfauV16}, providing one of the first practical applications of this important theoretical construct. In more detail, we solve the instance segmentation problem as agglomeration of image superpixels, with the agent predicting the weights of the edges in the superpixel region adjacency graph. Based on the predicted weights, the segmentation is obtained through (non-differentiable) graph partitioning. The segmented objects are evaluated by the critic, which learns to approximate the rewards based on object- and image-level reasoning (see Fig.~\ref{fig_overview}).

The main contributions of this work can be summarized as follows: (i) we formulate instance segmentation as a RL problem based on a stateless actor-critic setup, encapsulating the non-differentiable step of instance extraction into the environment and thus achieving end-to-end learning; (ii) we do \emph{not} use annotated images for supervision and instead exploit prior knowledge on instance appearance and morphology by tying the rewards to the conformity of the predicted objects to pre-defined rules and learning to approximate the (non-differentiable) reward function with the critic; (iii) we introduce a strategy for spatial decomposition of rewards based on fixed-sized subgraphs to enable localized supervision from combinations of object- and image-level rules. (iv) we demonstrate the feasibility of our approach on synthetic and real images and show an application to two important segmentation tasks in biology. In all experiments, our framework delivers excellent segmentations with no supervision other than high-level rules.

\section{Related work}

Reinforcement learning has so far not found significant adoption in the segmentation domain. The closest to our work are two methods in which RL has been introduced to learn a sequence of segmentation decision steps as a Markov Decision Process. In the actor critic framework of \cite{Araslanov:2019:ACIS}, the actor recurrently predicts one instance mask at a time based on the gradient provided by the critic. The training needs fully segmented images as supervision and the overall system, including an LSTM sub-network between the encoder and the decoder, is fairly complex. In \cite{RLClustering}, the individual decision steps correspond to merges of clusters while their sequence defines a hierarchical agglomeration process on a superpixel graph. The reward function is based on Rand index and thus not differentiable, but the overall framework requires full (super)pixelwise supervision for training.

Reward decomposition was introduced for multi agent RL by \cite{sunehag2017valuedecomposition} where a global reward is decomposed into a per agent reward. \cite{NIPS2005_02180771} proves that a stateless RL setup with decomposed rewards requires far less training samples than a RL setup with a global reward. In \cite{DBLP:journals/corr/abs-1910-08143} reward decomposition is applied both temporally and spatially for zero-shot inference on unseen environments by training on locally selected samples to learn the underlying physics of the environment. 

The restriction to differentiable losses is present in all application domains of deep learning. Common ways to address it are based on a soft relaxation of the loss that can be differentiated. The relaxation can be designed specifically for the loss, for example, Area-under-Curve \cite{eban2017scalable} for classification or Jaccard Index \cite{berman2018lovasz} for semantic segmentation. These approaches are not directly applicable to our use case as we aim to use a variety of object- and image-level priors, which should be combined without handcrafting an approximate loss for each case. More generally, but still for a concrete task loss, Direct Loss Minimization  has been proposed in \cite{song2016training}. For semi-supervised learning of a classification or ranking task, Discriminative Adversarial Networks have been proposed as a means to learn an approximation to the loss \cite{santos2017learning}. Most generally, \cite{grabocka2019learning} propose to train a surrogate neural network which will serve as a smooth approximation of the true loss. In our setup, the critic can informally be viewed as a surrogate network as it learns to approximate the priors through the rewards by Q-learning. 

Incorporation of rules and priors is particularly important in biomedical imaging applications, where such knowledge can be exploited to augment or even substitute scarce groundtruth annotations. For example, the shape prior is explicitly encoded in popular nuclear \cite{schmidt2018cell} and cellular \cite{stringer2021cellpose} segmentation algorithms based on spatial embedding learning. Learned non-linear representations of the shape are used in \cite{oktay2018anatomically}, while in \cite{hu_2019_topological} the loss for object boundary prediction is made topology-aware. Domain-specific priors can also be exploited in post-processing by graph partitioning \cite{pape2019leveraging}. Interestingly, the energy minimization procedure underlying the graph partitioning can also be incorporated into the learning step \cite{Maitin-Shepard2016combinatorial,song2019end,abbas2021combinatorial}.

\section{Methods}

\begin{figure*}
	\centering
	\includegraphics[width=0.7\textwidth]{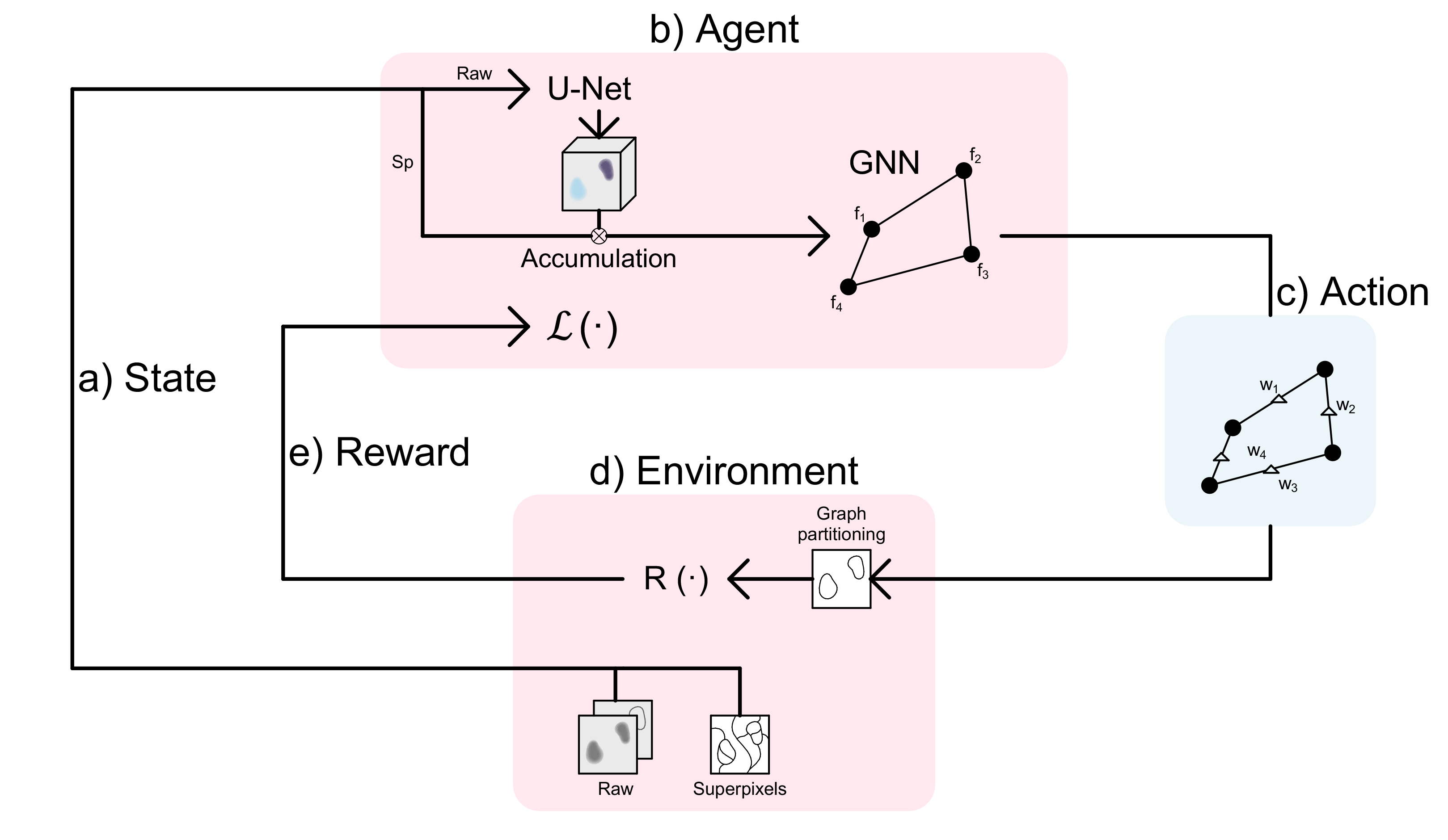}
	\caption{Interaction of the agent with the environment: (a) shows the state, which is composed of the image and superpixels; (b) depicts the agent, which consists of the actor and critic networks as well as the feature extractor that computes the node input features; (c) given the state, the agent performs the actions by predicting edge weights on the graph; (d) the environment, which includes the image, superpixels, graph and graph partitioning based on the weights predicted through agent actions ; (e) rewards are obtained by evaluating the segmentation arising from the graph partitioning, based on pre-defined and data dependent rules. The rewards are given back to the agent where they are used for training.
	}
	\label{fig_overview}
\end{figure*}

The task of instance segmentation can be formalized as transforming an image $x$ into a labeling $y$ that maps each pixel to a label value. 
An instance corresponds to the maximal set of pixels with the same label value.
Typically, the instance segmentation problem is solved via supervised learning, i.e. using a training set with groundtruth labels $\hat{y}$. Note that $y$ is invariant under the permutation of label values, which makes it difficult to formulate instance segmentation in a fully differentiable manner. Most approaches first predict a "soft" representation with a CNN, e.g. affinities \cite{lee2017superhuman,gao2019ssap,DBLP:journals/corr/abs-1904-12654}, boundaries \cite{beier2017multicut,funke2018large} or embeddings \cite{de2017semantic,neven2019instance} and apply non-differentiable post-processing, such as agglomeration \cite{funke2018large,bailoni2019generalized}, density clustering \cite{mcinnes2017accelerated,comaniciu2002mean} or partitioning \cite{andres2012globally}, to obtain the instance segmentation.
Alternatively, proposal-based methods predict a bounding-box per instance and then predict the instance mask for each bounding-box \cite{he2017mask}.
Furthermore, the common evaluation metrics for instance segmentation \cite{meilua2003comparing,rand1971objective} are also not differentiable.

Our main motivation to explore RL for the instance segmentation task is to circumvent the restriction to differentiable losses and - regardless of the loss - to make the whole pipeline end-to-end even in presence of non-differentiable steps that transform pixelwise CNN predictions into instances. 

We formulate the instance segmentation problem using a region adjacency graph $G = (V, E)$, where the nodes $V$ correspond to superpixels (clusters of pixels) and the edges $E$ connect nodes that belong to spatially adjacent superpixels.
Given edge weights $W$, the instance segmentation is obtained by partitioning the graph, here using an approximate multicut solver \cite{kernighan1970efficient}.
Together, the image data, superpixels, graph and the graph partitioning make up the environment $\mathcal{E}$ of our RL setup. Based on the state $s$ of $\mathcal{E}$, the agent $\mathcal{A}$ predicts actions $a$. Here, the actions are interpreted as edge weights $W$ and used to partition the graph.
The reward $r$ is then computed based on this partitioning.
Our agent $\mathcal{A}$ is a stateless actor-critic \cite{DBLP:journals/corr/abs-1812-05905}, represented by two graph neural networks (GNN) \cite{DBLP:journals/corr/GilmerSRVD17}.
The actor predicts the actions $a$ based on the graph and its node features $F$. The node(superpixel) features are computed by pooling together the corresponding pixel features based on the raw image data. 

We compute the node features $F$ with a UNet \cite{DBLP:journals/corr/RonnebergerFB15} that takes the image as input and outputs a feature vector per pixel. These features are spatially averaged over the superpixels to obtain $F$. The feature extractor UNet is part of the agent $\mathcal{A}$, thus training it end-to-end with the actor and critic networks (Fig.~\ref{fig_overview}). In low data regimes it is also possible to use a pre-trained and fixed feature extractor or to combine the learned features with hand-crafted ones.

Crucially, the reinforcement setup enables us to use both a non-differentiable instance segmentation step and reward function, by encapsulation of the ``pixels to instances'' step in the environment and learning a policy based on the rewards with the stateless actor critic.

\subsection{Stateless Reinforcement Learning Setup} \label{sec:stateless_rl}

Unlike most RL settings \cite{Sutton1998}, our approach does not require an explicitly time dependent state:
the actions returned by the agent correspond to the real-valued edge weights in $[0,1]$, which are used to compute the graph partitioning. Any state can be reached by a single step from the initial state and there exists no time dependency in the state transition. Unlike \cite{RLClustering}, we predict all edge values at once which allows us to avoid the iterative strategy of \cite{Araslanov:2019:ACIS} and deliver and evaluate a complete segmentation in every step.
Hence, we implement a stateless actor critic formulation. 

Stateless RL was introduced in \cite{DBLP:journals/corr/PfauV16} to study the connection between generative adversarial networks and actor critics, our method is one of the first practical applications of this concept.
Here, the agent consists of an actor, which predicts the actions $a$ and a critic, which predicts the action value $Q$ (expected future discounted reward) given the actions.
The stateless approach simplifies the action value: it estimates the reward for a single step instead of the expected sum of discounted future rewards for many steps. We have explored a multi-step setup as well, but found that it yields inferior results for our application; details can be found in the App.~\ref{sec:multistep}. Furthermore, we compute sub-graph rewards instead of relying on a single global reward in order to provide a more localized reward signal (see Section \ref{sec:subgraph_supervision} for details).

The actor corresponds to a single GNN, which predicts the mean and variance of a Normal distribution for each edge.
The actions $a$ are determined by sampling from this distribution and applying a sigmoid to the result to obtain continuous edge weights in the value range $[0,1]$.
The GNN takes the state $s=(G, F)$ as input arguments 
and its graph convolution for the $i^{th}$ node is defined as in \cite{DBLP:journals/corr/GilmerSRVD17}:
\begin{align} \label{eq:gnn}
f_i = \gamma_{\pi} \left(f_i, \frac{1}{\left| N(i) \right|} \sum_{j \in N(i)}  \phi_{\pi} \left(f_i, f_j \right) \right)
\end{align}
where $\gamma_{\pi}$ as well as $\phi_{\pi}$ are MLPs, $(\cdot, \cdot)$ is the concatenation of vectors and $N(i)$ is the set of neighbors of node $i$.
The gradient of the loss for the actor is given by:
\begin{equation} \label{eq:gradient_loss}  
	\nabla_\theta \mathcal{L}_{actor} = \nabla_\theta \frac{1}{\left|SG\right|} \sum_{sg \in G} \left[ \alpha \sum_{\hat{a}\in sg}log(\pi^\theta(\hat{a}|s)) - Q_{sg}(s, a) \right]
\end{equation}
This loss gradient is derived following \cite{DBLP:journals/corr/abs-1812-05905}. We adapt it to the sub-graph reward structure by calculating the joint action probability of the policy $\pi^\theta$ over each sub-graph $sg$ in the set of all sub-graphs $SG$. Using this loss to optimize the policy parameters $\theta$ minimizes the Kullback-Leibler divergence between the Gibbs distribution of action values for each sub-graph $Q_{sg}(s, a)$ and the policy with respect to the parameters $\theta$ of the policy. $\alpha$ is a trainable temperature parameter which is optimized following the method introduced by \cite{DBLP:journals/corr/abs-1812-05905}.\\

The critic predicts the action value $Q_{sg}$ for each sub-graph ${sg} \in SG$.
It consists of a GNN $Q_{sg}(s, a)$ that takes the state $s=(G, F)$ as well as the actions $a$ predicted by the actor as input and predicts a feature vector for each edge.
The graph convolution from Equation \ref{eq:gnn} is slightly modified:
\begin{align}
f_i = \gamma_{Q} \left(f_i, \frac{1}{\left| N(i) \right|} \sum_{j \in N(i)}  \phi_{Q} \left(f_i, f_j, a_{(i, j)} \right) \right)
\end{align}
again $\gamma_{Q}$ and $\phi_{Q}$ are MLPs. 
Based on these edge features $Q_{sg}$ is predicted for each sub-graph via an MLP. Here, we use a set of subgraph sizes (typically, 6, 12, 32, 128) to generate a supervison signal for different neighborhood scales. A given MLP is only valid for a fixed graph size, so we employ a different MLP for each size. 
The loss for the critic is given by:
\begin{equation} \label{eq:loss_critic} 
	\mathcal{L}_{critic} = \frac{1}{\left|SG\right|} \sum_{sg \in G}\frac{1}{2}(Q^\delta_{sg}(s, a) - r) ^ 2
\end{equation}
Minimizing this loss with respect to the action value function's parameters $\delta$ minimizes the difference between the expected reward and action values $Q^\delta_{sg}(s, a)$.

\subsection{Localized Sub-graph Rewards} \label{sec:subgraph_supervision}

In most RL applications a global scalar reward is provided per state transition.
In our application of graph-based instance segmentation, it is instead desirable to introduce several more localized rewards in order to learn from a reward for the specific action, rather than a global scalar. Here, reward decomposition is natural because we evaluate the segmentation quality per object and can use the object scores to provide a localized reward. In order to formalize this idea, we have designed our actor critic (Section \ref{sec:stateless_rl}) to learn from sub-graph rewards.

\begin{figure}
	\centering
	\begin{minipage}[b]{0.45\textwidth}
	    \centering
    	\includegraphics[width=1.\textwidth]{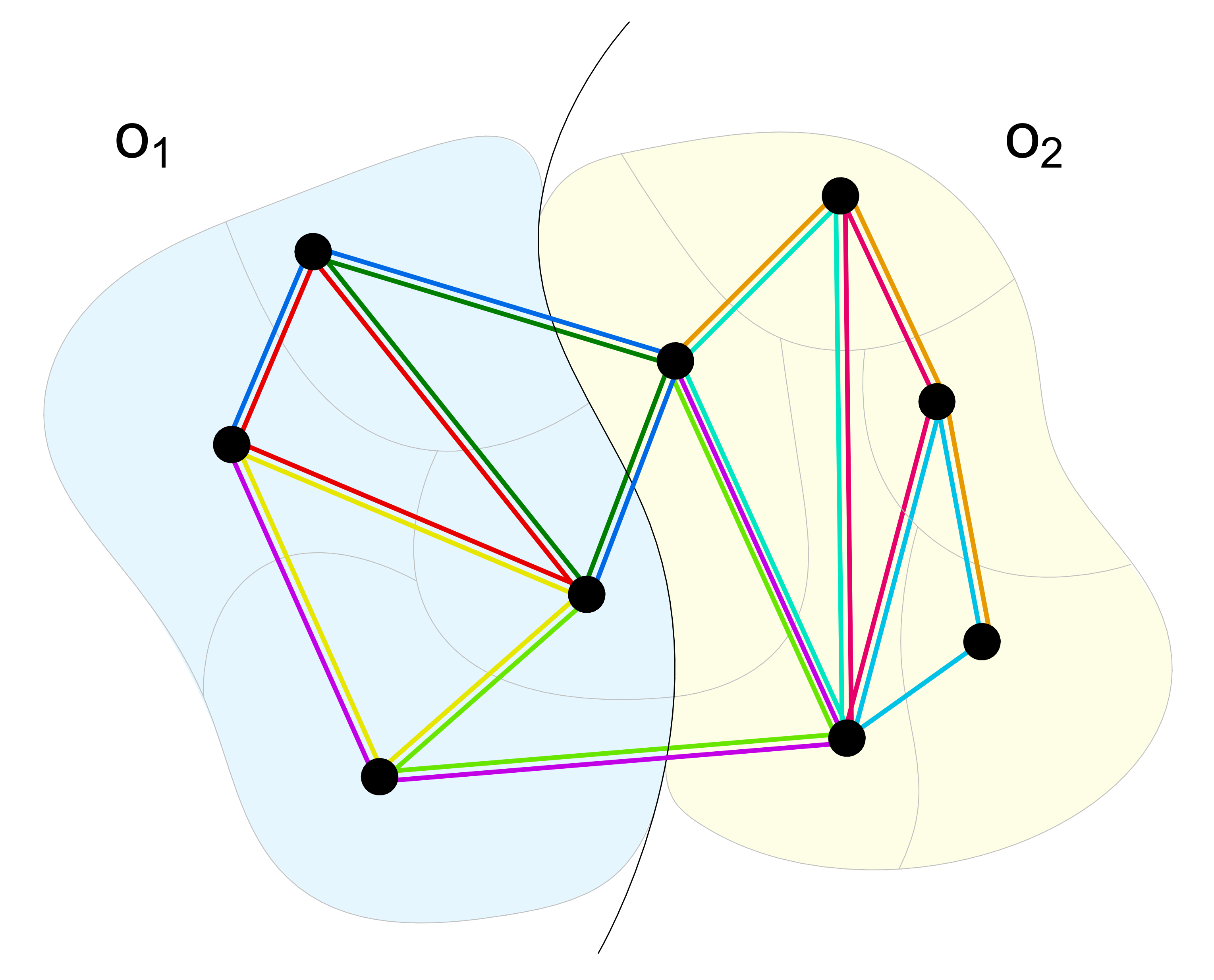}
    	\caption{The graph is subdivided into sub-graphs, each sub-graph is highlighted by a different color. All sub-graphs have the same number of edges (here 3). Overall, we use a variety of sizes covering different notions of locality.}
    	\label{fig_sg}
	\end{minipage}
    \hspace{.05\linewidth}
	\begin{minipage}[b]{0.45\textwidth}
	    \centering
    	\includegraphics[width=0.9\textwidth]{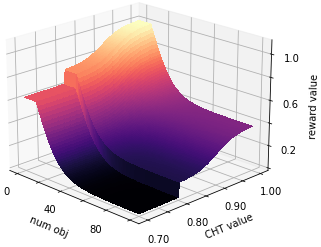}
    	\caption{An example reward landscape Circle Hough Transform (CHT) rewards. High rewards are given if the overall number of predicted objects is not too high and if the respective object has a large CHT value. }
    	\label{fig_reward_landscape}
	\end{minipage}
\end{figure}

A good set of sub-graphs should fulfill the following requirements: each sub-graph should be connected so that the input to the MLP that computes the activation value for the sub-graphs is correlated. The size of the sub-graphs should be adjustable and all sub-graphs should be extracted with the exact same size to be valid inputs for the MLP. The union of all sub-graphs should cover the complete graph so that each edge contributes to at least one action value $Q_{sg}$. The sub-graphs should overlap to provide a smooth sum of action values. We have designed Alg.~\ref{algo:sgs} to extract a set of sub-graphs according to these requirements. Fig.~\ref{fig_sg} shows an example sub-graph decomposition.

While some of the rewards used in our experiments can be directly defined for sub-graphs, most are instead defined per object (see App.~\ref{sec:rewards} for details on reward design). We use the following general procedure to map object-level rewards to sub-graphs: first assign to each superpixel the reward of its corresponding object. The reward per edge is determined by the maximum value of its two incident superpixels' rewards. The edge rewards are averaged to obtain the reward per sub-graph.

By taking the maximum we assign the higher score to edges whose incident superpixels belong to different objects, because they probably correspond to a correct split. 
Note that the uncertainty in the assignment of low rewards can lead to a noisy reward signal, but the averaging of the edge rewards over the sub-graphs and the overlaps between the sub-graphs smooth the rewards.
We have also explored a different actor critic setup that can use object level rewards directly, with no sub-graph extraction and mapping. However, this approach yields inferior results, see App.~\ref{sec:object_rewards} for details.

\section{Experiments} \label{sec:exp}

We evaluate our approach on three instance segmentation problems: one synthetic and two real. For a proof-of-principle, we construct a synthetic dataset with circular shapes on structured background, showing how our framework can exploit simple geometric priors.
Next, we apply the method to a popular microscopy benchmark dataset for nucleus segmentation \cite{caicedo2019nucleus}.
Finally, we consider a challenging biological segmentation problem with boundary-labeled cells. Here, we evaluate both learning restricted to prior rules and mixed supervision combining rule-based and direct rewards computed from groundtruth annotations.
The problem setup, network architectures and hyperparameters are reported in detail in App.~\ref{sec:exp_details}.

\subsection{Synthetic Data: Circles on Structured Ground}\label{sec_synth_data}
We create synthetic images of circles on a structured background and segment this data using only simple geometric rules. Superpixels were generated with the mutex watershed \cite{DBLP:journals/corr/abs-1904-12654} applied to the gaussian gradient of the image.
Here, we demonstrate that the actor critic can be trained without any direct object-level supervision and apply a simplified setup with a fixed pixel feature extractor, pre-trained through self-supervision (see App.~\ref{selfsup_pt}). 

The object-level reward is based on the Circle Hough Transform (CHT) \cite{DBLP:journals/corr/HassaneinMSR15}. It is combined with an estimate for the total number of objects in the image as an additional global reward.
The global reward gives useful gradients during early training stages: when too few potential objects are found in the prediction, a low reward can be given to the tentative background object. If too many potential objects are found, a low reward can be given to all the foreground objects with a low CHT value.
The surface created by the per-object and global reward is shown in Fig.~\ref{fig_reward_landscape}. The exact reward computation can be found in App.~\ref{sec:circle_rewards}.

\begin{figure}[!tbp]
    \centering
    \subfloat[Reinforcement learning output.]{\includegraphics[width=.5\textwidth]{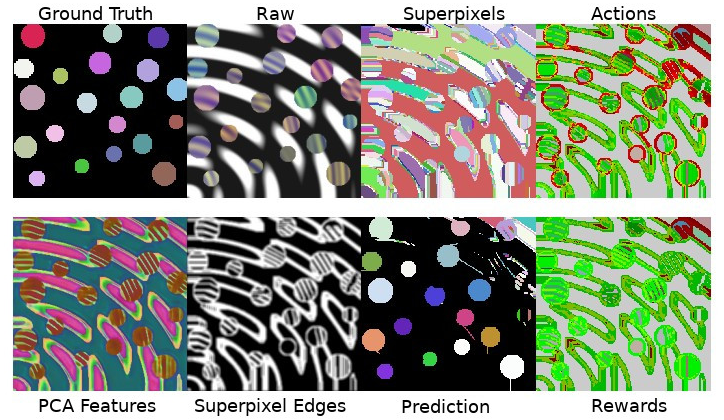}\label{fig_exp_crcl}}
    \hfill
    \subfloat[Mutex watershed baseline.]{\includegraphics[width=.3\textwidth]{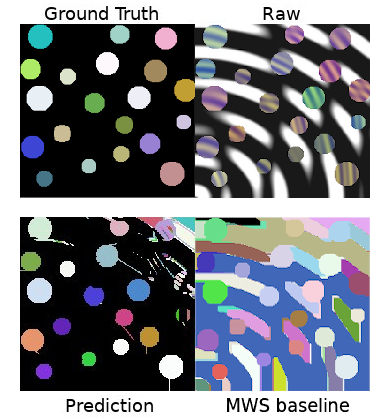}\label{fig_mws_crcl}}
	\caption{Synthetic data. a) top left to right: groundtruth segmentation, raw data, superpixels and visualization of  the actions (merge actions in green, split actions in red). Bottom left to right: pre-trained pixel embeddings, superpixel edges, segmentation result and visualization of the rewards (light green for high rewards, dark red for low rewards. b) comparison of segmentation from our method and the mutex watershed.}
	\label{fig_exp_mws_crcl}
	\vspace{-0.25cm}
\end{figure}

Fig.~\ref{fig_exp_mws_crcl} shows the output of all algorithm components on a sample image. We also computed results with the mutex watershed \cite{DBLP:journals/corr/abs-1904-12654}, a typical algorithm for boundary based instance segmentation in microscopy. Texture within objects and structured background are inherently difficult for region-growing algorithms, but our approach can exploit higher-level reasoning along with low-level information and achieve a good segmentation.

\subsection{Real Data: Nucleus Segmentation}
Nuclei are a very frequent target of instance segmentation in microscopy, which is also reflected in the large amount of publicly available annotated  data. The availability of training data sparked the development of popular pre-trained solutions, such as a generalist UNet \cite{falk2019u} or StarDist \cite{schmidt2018cell} and CellPose \cite{stringer2021cellpose} which both have an (implicit) shape prior. Also, due to ubiquity of nuclei in microscopy, detailed prior knowledge exists on their shape and their appearance under different stainings. The experiments in this section aim to answer the following questions: i) given fully annotated groundtruth images for training, is there an advantage in using our RL formulation with object-level rewards compared to commonly used fully supervised baselines? ii) given superpixels that can be combined into the correct solution, but no other direct supervision, can our approach learn to combine the superpixels correctly only from high-level rules? iii) what happens if superpixels are suboptimal? For data, we turn to the dataset of \cite{caicedo2019nucleus} and select images that contain nuclei of medium size (175 for training and 22 for test). 

\begin{wraptable}{r}{.47\textwidth}
\begin{tiny}
\centering
\begin{tabular}{lllll}
\toprule
       Method & Superpixel &  mAP & IoU50 & IoU75 \\
\midrule
         UNet & -   &0.710 & 0.900 & 0.756 \\
     StarDist & -   &0.645 & 0.938 & 0.736 \\
     Cellpose & -   &0.666 & 0.931 & 0.776 \\
  UNet + MC   & GT    & 0.674 & 0.806 & 0.702 \\
  ours (sup.) & GT  & \textbf{0.766} & 0.907 & 0.799 \\
\midrule
         Otsu & -    &  0.554   & 0.763 & 0.579 \\
ours (unsup.) & GT   & \textbf{0.743} & 0.916 & 0.787 \\
ours (unsup.) & UNET & 0.671 & 0.872  & 0.704  \\
ours (unsup.) & RAW  & 0.453 & 0.785  & 0.439 \\
\midrule
sp gt  & UNET &  0.793 & 0.98 & 0.852 \\
sp gt  & RAW  &  0.554 & 0.969 & 0.505 \\
\bottomrule
\end{tabular}
\caption{Nuclei segmentation: for mAP and IoU higher values are better. Methods above the first middle line were trained fully supervised. Methods below the first middle line were trained without groundtruth, the results below the second middle line indicate the quality of the superpixels projected to the groundtruth (best possible result that can be achieved with the given superpixels).}
\label{tab:dsb_results}
\end{tiny}
\end{wraptable}

Features are learned end-to-end. In the unsupervised setting, we compute the reward by combining several object descriptors: eccentricity, extent, minor diameter, perimeter, solidity as well as mean, maximum and minimum intensity per object. The object reward is then given by the normalized sum of square distances of these quantities and their expected value. Objects larger than 15,000 pixels are considered to belong to the background and are not assigned a reward.
Since the superpixels serve as fixed input into our model that do not get modified, the accuracy of our segmentations is bound by their accuracy. To investigate their influence, we evaluate our approach with three different sets of superpixels: ``GT'', where we  intersect the superpixels with the groundtruth object masks to ensure that a correct segmentation can be recovered, ``UNET'', where we compute the superpixels using predictions of a pre-trained U-Net as an edge detector and ``RAW'', where we only take into account the raw image data. See \ref{sec:dsb} for more details on superpixels and object descriptors.

Tab.~\ref{tab:dsb_results} summarizes the results, with a comparison to popular generalist pre-trained nuclear segmentation methods: StarDist \cite{schmidt2018cell}, Cellpose \cite{stringer2021cellpose} and UNet \cite{falk2019u}. For StarDist and Cellpose, we use the pre-trained models provided with the papers. The UNet is trained on the same images as StarDist, the instance segmentation is recovered either by applying connected components to the boundary-subtracted foreground prediction (``UNet'') or, to obtain a comparison conditioned on a particular set of superpixels, by using the UNet boundary predictions and superpixels described above as input to Multicut graph-based agglomeration (``UNet + MC''). Otsu threshold serves as a simple unsupervised baseline \cite{otsu1979threshold}, where binarizing the image is followed by connected components to obtain the instance segmentation.

For the first question, we train our pipeline fully supervised (``ours (sup.)'') as described in App.~\ref{sec:direct_superivision}: we use pixelwise groundtruth, but can also exploit our RL formulation where the loss is assigned to individual objects through the non-differentiable graph agglomeration step. Here, our method performs better than all baselines without RL, so there is clearly an advantage to using object-level supervision (as also demonstrated recently for non-RL setups, e.g. by \cite{wolny2021sparse}). 

For the other two questions, we train our method using only rule-based rewards (``ours (unsup.)''). Given superpixels from which the groundtruth image can be recovered (''GT''), we then achieve better segmentation quality than the fully supervised baselines and the gap in performance between our unsupervised and supervised approach is smaller than the gap to the runner-up baseline. Of note, our unsupervised model also outperforms the ``UNet + MC'' baseline using the same ''GT'' superpixels, so its performance cannot be explained just by the use of groundtruth in superpixel generation. Example results and failure cases are shown in App.~\ref{sec:dsb}. 

In the third experiment, we use a pretrained UNet as an edge detector to create superpixels of ''medium'' quality and again obtain strong results, outperforming StarDist, CellPose and UNet+MC with ''GT'' superpixels. Finally, with our worst set of superpixels obtained directly from the raw data, the method can learn to exploit the rules, but is clearly hindered by the superpixel quality.

\subsection{Real Data: Cell Segmentation} \label{sec_exp_bio}
Biomedical applications often require segmentation of objects of known morphology arranged in a regular pattern \cite{thompson_1992}.
Such data presents the best use case for our algorithm as the reward function can leverage such priors.
We address a cell segmentation problem from developmental biology, where cells often follow stereotypical shapes and arrangement: 317 drosophila embryo images from \cite{bhide2020semiautomatic}, including 10 with expert annotations used as test data.
Note that several pre-trained networks are available for cell segmentation in light microscopy \cite{stringer2021cellpose,vonChamier2021,wolny2020accurate}; however, they produce sub-par results on this data due to experimental differences.

\begin{figure*}
	\centering
	\includegraphics[width=0.85\textwidth]{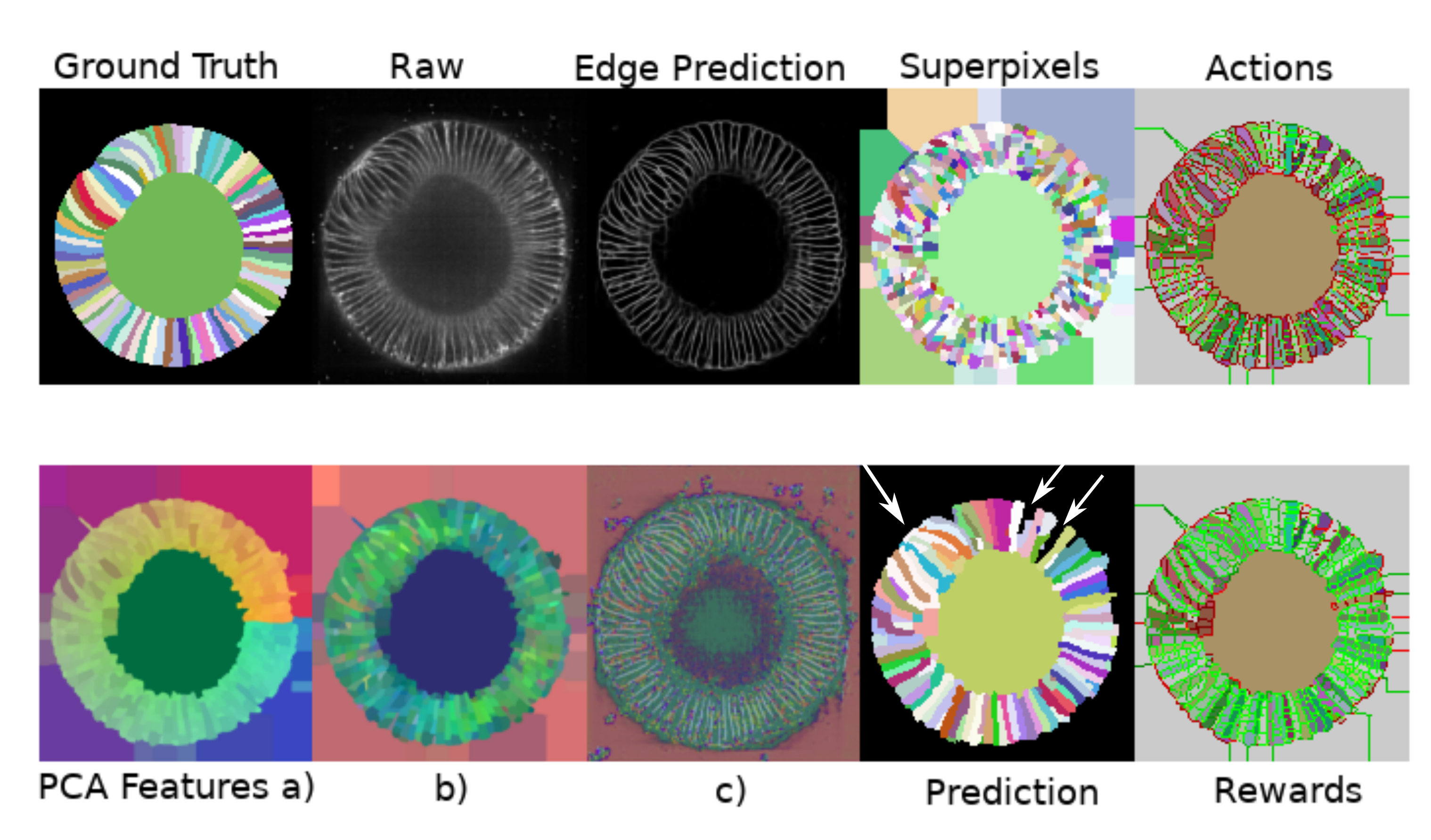}
	\caption{Cell segmentation experiment. Top left to right: groundtruth segmentation; raw data; boundary predictions; superpixel over-segmentation; visualization for the actions on every edge (green = merge action, red = split action). Bottom left to right: a) handcrafted features per superpixel; b) learned features averaged over superpixels; c) learned features per pixel; Multicut segmentations; visualization of the rewards (light green = high reward, dark red = low reward). For all features, we use the first 3 PCA components for visualisation. White arrows point to remaining errors.}
	\label{fig_sample_leptin}
\end{figure*}

\begin{wraptable}{r}{.47\textwidth}
\begin{tiny}
    \centering
    \begin{tabular}{llll}
        \toprule
        Method                                   & VI           & VI merge      & VI split \\
        \midrule
        sp gt                                    & 1.266 & 0.672 & 0.594              \\
        \cmidrule(r){1-4}
        ours                                     & 2.213 & 0.839 & 1.374              \\
        ours (semisup.)                          & \textbf{1.634} & 0.733 & 0.901              \\  
        ours (handcrafted)                       & 2.523 & 0.987 & 1.536              \\
        UNet + MC                                & 3.361 & 3.019 & 0.342              \\
        contrastive                              & 4.440 & 1.155  & 3.28             \\
    \end{tabular}
    \caption{Cell segmentation results, measured by variation of information (VI) \cite{meilua2003comparing}. This entropy-based metric is commonly used to evaluate crowded segmentations in microscopy. We also report its merge and split component that measure the over-/under-segmentation error respectively. Lower values are better.}
   \label{leptin_results}
\end{tiny}
\end{wraptable}
The rewards express that the cells are arranged in a radial pattern, with the average size known from other experiments (see Fig.~\ref{fig_sample_leptin}). We set a high reward for merging superpixels that certainly belong to the background (close to the image boundary or center). For background edges near the foreground area, we modulate the reward by the circularity of the overall foreground contour. For the likely foreground edges, we compute object-level rewards by fitting a rotated bounding box to each object and comparing its radii and orientation to template values. We use a weight profile based on the known embryo width to combine object and background rewards (App.~\ref{sec:gaussian_weight}).

More formally, the rewards are calculated as follows: for each edge, we define the position $h$ as the average of the centers of the two incident superpixels. 
Given the image center $c$, the radius of a circle that approximately covers the foreground $j$ and the (maximal) image border position $m$, we use a gaussian kernel $\mathcal{K}(\cdot)$ for weighting and define edge reward $r_{edge}$:

\begin{wrapfigure}{l}{.5\textwidth}
\begin{align} \label{eq:reward_leptin}
    r_{bg} &=     
    \begin{cases}
        \mathcal{K}\left(\frac{||h-c||}{\gamma}\right) (1-a),& \text{if } h\leq j\\
        \mathcal{K}\left(\frac{||m-h||}{\eta}\right) (1-a),              & \text{otw}
    \end{cases}\\
    r_{fg} &= \mathcal{K}\left(\frac{||h-j||}{\delta}\right) \textrm{max}(r_{o1}, r_{o2}) \\
    r_{edge} &= r_{fg} + r_{bg}
\end{align}
\end{wrapfigure}
Here $\gamma$, $\eta$ and $\delta$ are normalization constants. The kernel function in Eq.~\ref{eq:reward_leptin} determines the background probability of an edge; $1-a$ constitutes a reward that favors merges. It is scaled by the background probability. 
The object rewards $r_{o}$ are found by fitting a rotated bounding box to the object and then comparing orientation and extent to expected values known from previous experiments. They are mapped to edge rewards $r_{o1}, r_{o2}$ using the maximum value of the two incident objects.

We pre-compute superpixels by using boundary predictions as input to a watershed seeded from local maxima. We use the UNet from \cite{wolny2020accurate}, which was trained on roughly similar images. As it was trained on plant cells in different microscope modality, its prediction is far from perfect, especially around the inner circle, see Fig.~\ref{fig_sample_leptin}``Edge prediction''.
We combine the learned node features with hand-crafted features:
the normalized polar coordinate of the superpixel center and the normalized superpixel size. Fig.~\ref{fig_sample_leptin} shows visualisations of the learned and hand-crafted features.
Interestingly, the learned features converge to a representation that resembles a semantic segmentation of boundaries.

Tab.~\ref{leptin_results} shows the results: ``ours'' is the method described above; for ``ours (semisup.)'' we train a model that additionally receives direct supervision from groundtruth for a single patch using the reward from App.~\ref{sec:direct_superivision} and for ``ours (handcrafted)'' we only use the hand-crafted features and not the learned features.
We include the UNet from \cite{wolny2020accurate} with Multicut for instance segmentation (``UNet + MC'') as well as the method of \cite{de2017semantic} trained on the same data as \cite{wolny2020accurate} (``contrastive'') as baselines. Since only 10 images of the dataset are annotated, we cannot efficiently finetune any of the popular cell segmentation networks on this dataset.
We also project the superpixels to their respective groundtruth cluster (``sp gt'') to indicate the best possible solution that can be achieved with the given superpixels.
Our approach clearly outperforms the baseline methods trained on the data from \cite{wolny2020accurate}. While predictions are not perfect (white arrows in Fig.~\ref{fig_sample_leptin}, prior rules turn out to be sufficient to assemble most cells correctly. The remaining errors are caused by objects not fully conforming to the priors ("bent" rather than straight oval cells) or by a very weak boundary prediction.
Furthermore, we see that the learned features significantly improve results and that the semi-supervised approach provides a large boost, even with a single patch used for direct supervision.
We only report results for the best model as measured by the reward on a validation set across several training runs. App.~\ Fig.~\ref{fig_reward_evo} shows validation reward curves consistently improve during training for all random seeds.

\vspace{-0.45cm}
\section{Discussion and Outlook}
\vspace{-0.35cm}
We introduced an end-to-end instance segmentation algorithm that can exploit non-differentiable loss functions and high-level prior information. Our novel RL approach is based on the stateless actor-critic and predicts the full segmentation at every step, allowing us to assign rewards to all objects and reach stable convergence. The segmentation problem is formulated as graph partitioning; we design a reward decomposition algorithm which maps object- and image-level rewards to sub-graphs for localized supervision.
Our experiments demonstrate good segmentation quality on synthetic and real data using \emph{only} rule-based supervision without any object- or pixel-level labels, such as centers, boxes or masks. Furthermore, in case of full supervision, our method enables end-to-end instance segmentation with direct object-level reasoning, which will allow for post-processing-aware training of segmentation CNNs. 
In the future, we plan to explore other tasks and reward functions and will further study the semi-supervised setup that showed very promising initial results. 

\vspace{-0.2cm}
\paragraph{Limitations} Our method relies on superpixels which are fixed and not optimized jointly, so the upper bound on the performance is defined by superpixel accuracy. We believe an extension to pixels is possible and envision working on this in the future, but the current setup will not scale to the pixel level directly. Also, our method is limited to problems where consistent prior rules can be formulated for all instances. While this is the case for many applications in life science and medical imaging, not all object classes in the natural world can be described in this way. Here, our method could contribute by complementing training annotations with rules, reducing the overall labelling load in a semi-supervised setting. Finally, our approach requires non-trivial reward engineering as a trade-off for not performing any annotations.

{
    \small
    \bibliography{main}
    \bibliographystyle{iclr2023_conference}

}

\clearpage
\appendix
\section{Appendix}

\input{appendix}

\end{document}

%% file: math_commands.tex
\usepackage{amsmath,amsfonts,bm}

\def\eqref#1{equation~\ref{#1}}

\def\1{\bm{1}}

\DeclareMathAlphabet{\mathsfit}{\encodingdefault}{\sfdefault}{m}{sl}
\SetMathAlphabet{\mathsfit}{bold}{\encodingdefault}{\sfdefault}{bx}{n}



%% file: appendix.tex






\appendix
\section{Appendix}

\subsection{Self-supervised pretraining} \label{selfsup_pt}

\begin{figure*}
	\centering
	\includegraphics[width=.8\textwidth]{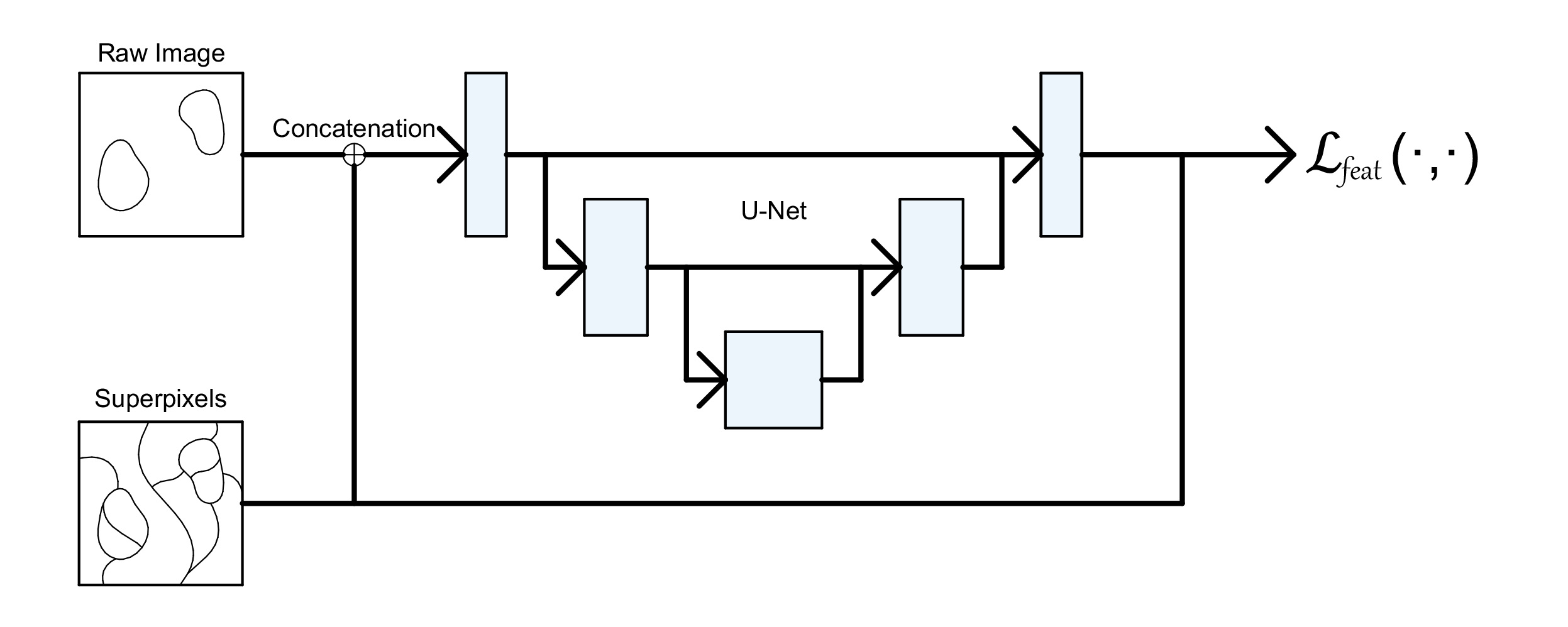}
	\caption{Training setup of the feature extractor. The input is a concatenation of the raw data and a smoothed edge map of the superpixels. The superpixel over-segmentation is used in the loss again as the supervision for learning the embedding space.}
	\label{fig_fe_sketch}
\end{figure*}

For self-supervised pre-training, we use a method based on the contrastive loss formulation of \cite{brab2017semantic}.
Consider a graph $G = (V, E)$, where the nodes in $V = \{1, 2, ..., n\}$ correspond to the individual superpixels and the edges in $E = \{(i, j) | i\neq j \, \textrm{and} \, i, j \in V \}$ connect nodes with adjacent superpixels.
In addition, consider edge weights $W \in \mathbb{R}^{|E|}$ associated with each edge. Here, we infer the weights from pixel-wise boundary probability predictions and normalize the weights such that $\sum_{w \in W} w = 1$ holds.
We train a 2D UNet to predict embeddings for each node in $V$ by pulling together pixel embeddings that belong to the same superpixel and pushing apart pixel embeddings for \emph{adjacent} superpixels.
The intensity of the push force is scaled by the weight of the respective edge.
With pixel embeddings $x_n$ and node embeddings $f_i = \frac{1}{m_i} \sum_{k \in s_i} x_k$, where $m_i$ is the mass of the superpixel for node $i$ and $s_i$ is the set of indices for all pixels of the corresponding superpixel,
and in accordance with \cite{brab2017semantic} we formulate the loss as
\begin{align}
	\mathcal{L}_{var} &= \frac{1}{|N|} \sum_{i=1}^{|N|} \frac{1}{m_i} \sum_{n=1}^{m_i} \left[ \text{\emph{d}}(f_i, x_n) - \delta_v \right]_+^2 \\
	\mathcal{L}_{dist} &= \sum_{(i, j)\in E}^{|E|} w_{(i, j)} \left[ 2\delta_d - \text{\emph{d}}(f_i, f_j) \right]_+^2 \\
	\mathcal{L}_{feat} &= \mathcal{L}_{var} + \mathcal{L}_{dist}
\end{align}
Here $\left[ \cdot \right]_+$ refers to selecting the max value from the argument and $0$. The forces are hinged by the distance limits $\delta_{var}$ and $\delta_{dist}$. $\text{\emph{d}}( \cdot )$ refers to the distance function in the embedding space. Since the feature extractor is trained self-supervised, we give it a smooth edge map of the superpixels as well as the raw data as an input, see Fig.~\ref{fig_fe_sketch}.\\
The training of the feature extractor happens prior to training the agent for \textit{Method 2}.

\subsection{Reward Generation} \label{sec:rewards}

We seek to express the rewards based on prior rules derived from topology, shape, texture, etc. Rules are typically formulated per-object. Section \ref{sec:subgraphs} describes the object-to-sub-graph reward mapping. The reward function is part of the environment and the critic learns to approximate it via Q-learning, enabling the use of non-differentiable functions.

This approach can also be extended to semantic instance segmentation where in addition to the instance labeling a semantic label is to be predicted.
To this end, each predicted object is softly assigned to one of the possible classes and the reward is generated specifically for the predicted class.
We make use of this extension by separating the objects into a foreground and background class in our experiments.

In addition to the sub-graph rewards our approach can also be extended to global rewards by global pooling of the output of the critic GNN and adding the squared difference of global action value and reward to Equation \ref{eq:loss_critic}.
Alternatively, the global reward can be distributed onto the sub-graph rewards via a weighted sum of sub-graph reward and global reward.
In the second approach a different global reward can be specified per class in the case of the semantic instance segmentation formulation.
We make use of the per class global reward to encode a reward for the correct number of predicted objects in the experiments for synthetic data (Subesection \ref{sec_synth_data}).

The biggest challenge in designing the reward function is to avoid local optima. 
Since the reward is derived from each predicted object, we define the reward by extracting shape features, position, orientation and size of objects and compare them with our expectation of the true object's features. 
This similarity score should be monotonically increasing as the objects fit our expectation better. 
All used similarity functions are to a certain extend linear, however an exponential reward function can speed up learning significantly.
Consider an object level reward $r \in [0, 1]$, which is linear. We calculate the exponential reward by 
\begin{align}
    r_{exp}(r) = \frac{exp(r \theta)}{exp(\theta)}
\end{align}
where the factor $\theta$ determines the range of the gradient in the output.
We also find that it is better to compute the reward as a "distance function" of all relevant features rather than decomposing it into the features and simply summing up the corresponding rewards.
In our experiments the latter approach behaved quite unpredictably and often generated local optima which the agent could not escape.

\subsection{Object level rewards} \label{sec:object_rewards}

\begin{figure}
	\centering
	\includegraphics[width=.5\textwidth]{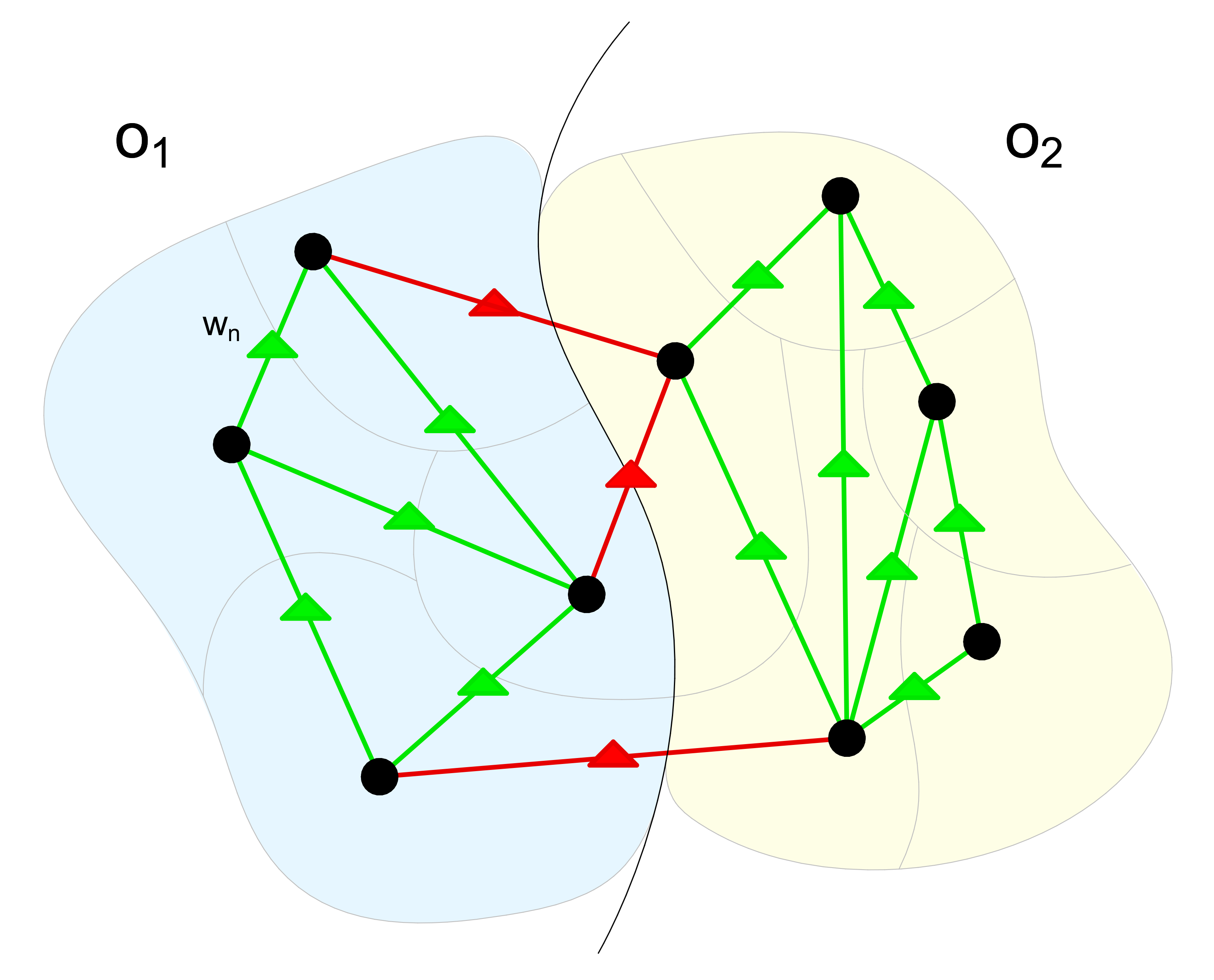}
	\caption{Object level rewards. We accumulate edge rewards over each object where we consider all edges that have at least one node within the respective object. E.g. for $o_1$ we consider all edges that are covered by the light blue object as well as all the red "split" edges.}
	\label{fig_obj_rew}
\end{figure}

We have tested generating the rewards based directly on the object scores instead of using the subgraph decomposition described in Section \ref{sec:subgraphs}.
Since rewards are mainly derived from the features of the predicted objects it seems reasonable to formulate the supervision signal directly for objects.
To this end we calculate a scalar reward per object as sketched in Figure \ref{fig_obj_rew}.
In this case, the critic uses a second GNN to predict the per-object action values. It is applied to an object's subgraph, which is composed of all edges that have at least one node in common with the respective object. 
The graph convolutions are followed by a global pooling operation which yields the scalar action value.
This GNN replaces the MLPs used in the case of the reward subgraph decomposition.
After extensive testing, we found that this approach is always inferior to the subgraph decomposition.

\subsection{Impact of different feature space capacities}
In Tab~\ref{table1} we compare the performance of our method, using different dimensionalities of the learned feature space (the number of channels in the output of the feature extractor UNet). We find that the reduced capacity of small feature spaces improves the agents performance. Here we train and evaluate on the fruit fly embryo dataset from Subsection \ref{sec_exp_bio}.

\begin{table}[!htbp]
    \centering
    \begin{footnotesize}
    \begin{tabular}{llll}
        \toprule
        n channels & VI      & VI merge & VI split \\
        \midrule
            4      & $1.712$ & $0.889$	& $0.823$ \\
            12	   & $2.062$ & $0.821$	& $1.241$ \\
            16     & $2.212$ & $0.838$	& $1.374$ \\
        \bottomrule
    \end{tabular}
    \caption{Quantitative evaluation of our method using different feature space dimensionality. We use the same metrics for evaluation as in Tab.\ref{leptin_results} and compare all results on the validation set.}
    \label{table1}
    \end{footnotesize}
\end{table}

\subsection{Random seed evaluation} \label{sec:random_seeds}
\begin{figure}[!htbp]
    \centering
    \subfloat[Setup 1. Features of size 16. 10 seeds.]{\includegraphics[width=.5\textwidth]{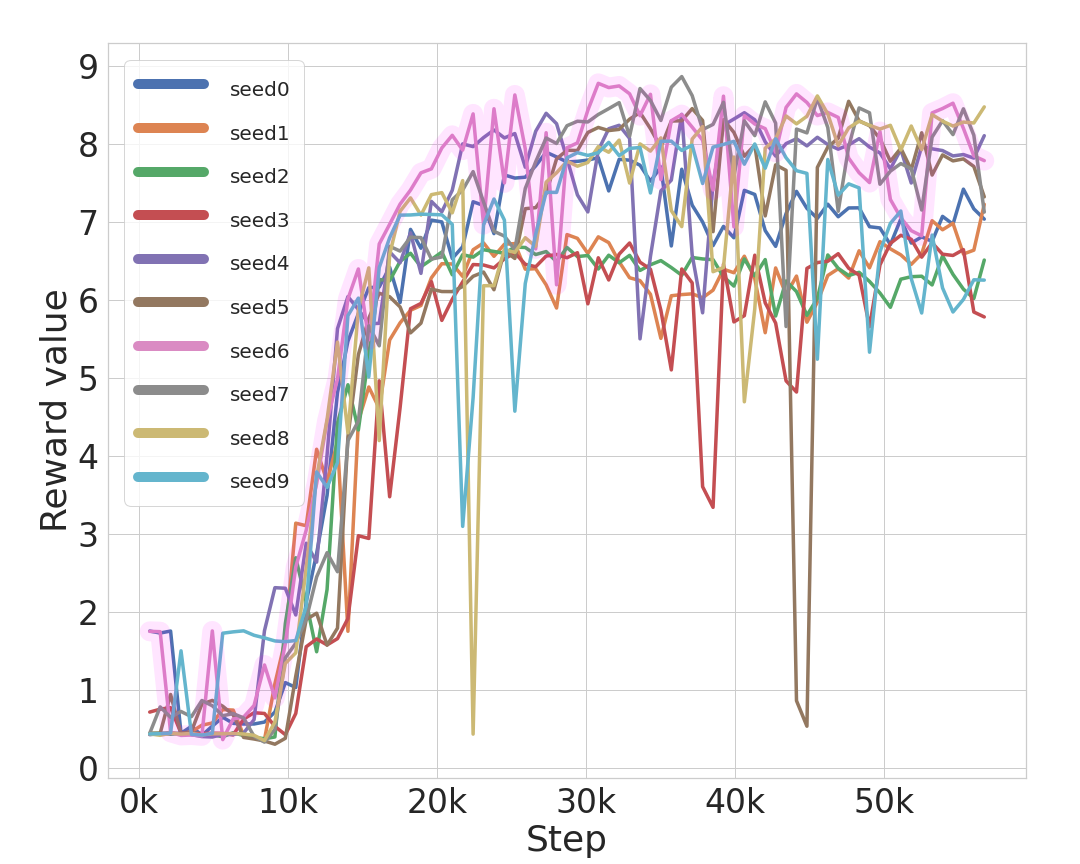}\label{fig:f1}}
    \hfill
    \subfloat[Setup 2. Features of size 12. 8 seeds.]{\includegraphics[width=.5\textwidth]{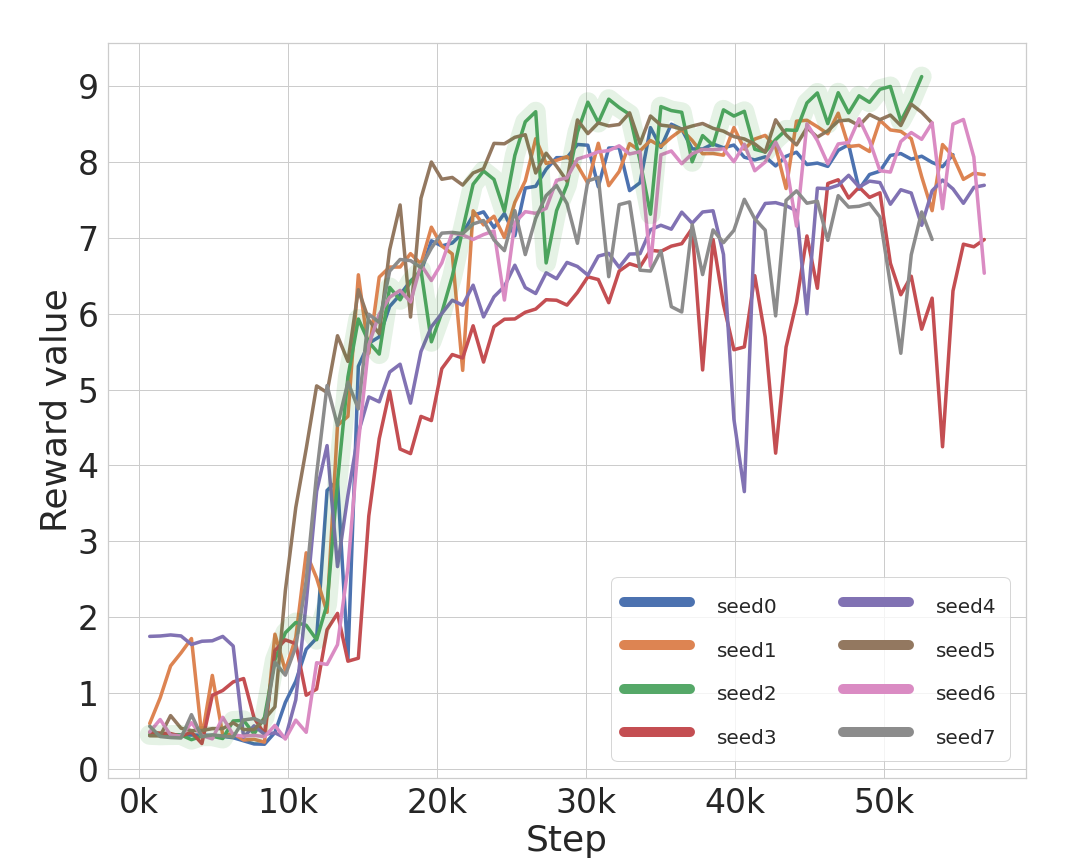}\label{fig:f2}}
    \caption{Running the same setup for different random seeds reveals a stable trend towards larger rewards. We select the model for comparison based on the best achieved reward (magenta line in Fig. \ref{fig:f1} and green line in Fig. \ref{fig:f2}).}
    \label{fig_reward_evo}
\end{figure}

Figure \ref{fig_reward_evo} shows the evolution of the average subgraph reward furing training for different random seeds. The model performance depends on the chosen seed and for the final comparisons we select the runs based on the best score. The seed is generated randomly for each run.

\subsection{Gaussian weighting scheme} \label{sec:gaussian_weight}
\begin{figure*}[!htbp]
    \centering
    \includegraphics[width=0.9\textwidth]{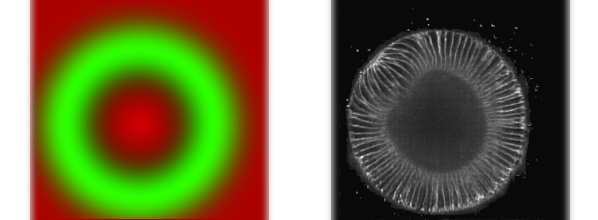}
    \caption{Weighting scheme for object rewards and merge affinity rewards, roughly encoding foreground location in \ref{sec_exp_bio}. Left: weights for object rewards in green and for merge affinity rewards in red, both have a gaussian profile and are concentric. Right: corresponding light-sheet image. }
    \label{gaussian_weighting}
\end{figure*}

Fig.~\ref{gaussian_weighting} shows the weighting scheme which was used to generate the rewards for the fruitfly embryo data (Subesection \ref{sec_exp_bio}). It can be seen as a very approximate semantic segmentation and serves the purpose of generating a reward maximum at the approximate foreground locations.

\subsection{Randomly generated subgraphs} \label{sec:subgraphs}
We select subgraphs using Alg.~\ref{algo:sgs}. Subgraphs are selected randomly starting from random nodes and continuously adding edges to the subgraph until the desired size is reached. The size of the subgraph is defined by the number of edges in the graph. The algorithm selects edges such that the subgraphs are connected and such that their density is high (low number of nodes in the subgraph).
\begin{algorithm}[htbp]
	\KwData{$G=(V, E)$, $l$}
	\KwResult{subgraphs by sets of $l$ edges}
	Initialization:$SG = \emptyset$\;
	\While{$E\backslash SG \neq \emptyset$}{
		pq = PriorityQueue\;
		prio = 0\;
		n\_draws = 0\;
		$sg = \emptyset$\;
		$sg_{vtx} = \emptyset$\;
		$i, j = (ij)$ s.t. $(ij)\in E\backslash SG$\;
		pq.push($i$, prio)\;
		pq.push($j$, prio)\;
		$sg = sg \cup (ij)$\;
		$sg_{vtx} = sg_{vtx} \cup i$\;
		$sg_{vtx} = sg_{vtx} \cup j$\;
		\While{|sg| < $l$}{
			$n$, n\_prio = pq.pop()\;
			n\_draws ++\;
			$adj = \{(nj) | \exists (nj)\in E \text{ and } \exists j \in sg_{vtx}\}$\;
			\ForAll{$(nj)\in adj$}{
				$sg = sg \cup (nj)$\;
				n\_draws = 0\;
			}
		\uIf{$|adj| < $ deg$(n)$}{
			n\_prio -= $(|adj|-1)$\;
			pq.push($n$, n\_prio)\;
		}
		\uIf{pq.size() $\leq$ n\_draws \& $\exists j |(nj) \in E, j\not\in sg_{vtx}$}{
			$j \in \left\{j |(nj) \in E, j\not\in sg_{vtx}\right\}$\;
			prio ++\;
			pq.push($j$, prio)\;
			$sg = sg \cup (nj)$)\;
			$sg_{vtx} = sg_{vtx} \cup j$\;
		}
		}
		$SG$ = $sg \cup SG$
	}
	\Return $SG$
	\caption{Dense subgraphs in a rag}
	\label{algo:sgs}
\end{algorithm}

\subsection{Multistep Reinforcement Learning} \label{sec:multistep}
We tested several methods that use multiple steps within one episode. In this formulation we predict the changes starting from an initial state rather than predicting absolute values for the edge weights.
For example, we can start from a state defined by edge weights derived from a boundary map.
Given that this state should be somewhat close to the desired state we expect that a few small steps within one episode should be sufficient.
In our experiments, we have typically used three steps per episode and used actions that can change the weight per edge by the values in $[-0.1, 0.1]$.
This approach generates an action space that is exponentially larger than in the stateless formulation.
A priori this setup might still be more stable because it is not possible to diverge from a given solution so fast due to the incremental changes per step.

Let us first consider the case with groundtruth edge weights.
In this case, we can give an accurate reward not only for the final segmentation but for every step. Hence, the path to the optimal state is linear.
Take for example an initial edge with weight $0.3$ and its respective ground truth edge with weight $0$. We can give a reward that mirrors the correct confidence of the action by using the negative value of the predicted action: $r=-a$.
This allows us to set the discount factor $\gamma$ of the RL setup to $0$, because the path to the correct edge weight is linear and the correct direction will be encoded in the reward at every step. 
Therefore the rewards for the following steps are not needed.
Setting the discount to $0$ generates a problem of equal size as the stateless RL method. However, for this approach the ground truth direction of the path must be known for each edge, so it can only be used in the case of full supervision.

To generalize the multistep approach to the rule-based rewards we need to choose a different setup where a constant reward is given at each non terminal step and the rule-based one is given at the terminal step.
This setup requires a discount factor $\gamma > 0$ and has an action space more complex than the stateless approach, because future steps are necessary to compute the reward. 
We tested this setup extensively against the stateless approach and found that it was not competitive.

\subsection{Network architectures, hyperparameters and experiment details} \label{sec:exp_details}

The U-Net used for feature extraction is based on a slightly modified implementation of the standard 2D U-Net. It uses max-pooling to spatially downsample the features by a factor of 2 in the encoder and transposed convolutions to upsample them in the decoder. All convolutions have a 3x3 kernel size, followed by ReLU activations and group normalization. It has four levels, each level consisting of 2 convolutional blocks followed by downsampling (encoder) / upsampling (decoder), the features from the encoder are concatenated to the decoder inputs on the same level using skip connections. For the experiments we use feature maps of size 32, 64, 128, 256 for the different levels; except for the nucleus segmentation experiment where we use 16, 32, 64, 128.

Both actor and critic are based on GNNS: the GNN for the actor predicts the action per edge and the critic has a GNN that also predicts per edge outputs. Both GNNs use an architecture of depth two and each layer consists of an MLP with three hidden layers and 2028 units per hidden layer. In the case of the critic we have an additional MLP per sub-graph size that takes the edge outpus of the GNN as input and predicts the action value for a given sub-graph. Each of these MLPs has two hidden layers with 2028 hidden units each.

We use the Adam optimizer with a learning rate of 0.0001 and the pytorch default values for all other hyperparameters. We have trained the networks on different GPUs, depending on the problem size (which is defined by the number of superpixels): Nvidia Geforce GTX 2080 TIs for the synthetic data experiments, Nvidia Geforce RTX 3090s for the nucleus segmentation experiments and Nvidia A100s for the cell segmentation experiments. A single training run always used a single GPU. 

The cell segmentation dataset is available on a general-purpose open-access repository Zenodo \footnote{\url{https://zenodo.org/record/4899944\#.YORWq0xRVEZ}}.

\subsection{Direct supervision} \label{sec:direct_superivision}
\begin{figure*}
	\centering
	\includegraphics[width=.7\textwidth]{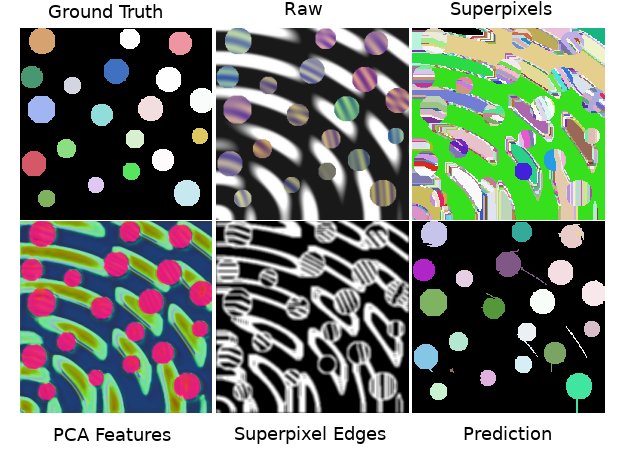}
	\caption{An example of a fully supervised prediction on the synthetic dataset. Here, we use the edge-based Dice score over subgraphs and make use of \textit{Method 1}, i.e. joint training of the feature extractor, but we initialized it using the weights pretrained by self-supervision. The features for the circles are significantly more pronounced after training.}
	\label{fig_full_sv}
\end{figure*}

\begin{figure*}
	\centering
	\includegraphics[width=1.\textwidth]{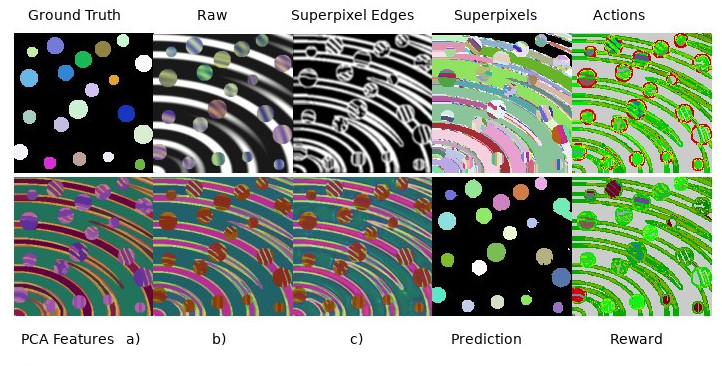}
	\caption{An example for  prediction with mixed supervision on the synthetic dataset. The reward was defined as follows: for all but one image we use the unsupervised CHT reward and for one image we make use of ground truth and the Dice score. We find that this mixed reward setting leads to improved performance compared to the unsupervised CHT reward.}
	\label{fig_mixed_sv}
\end{figure*}

We have also implemented a set-up for direct supervision that can be applied if any of the images have a groundtruth pixelwise segmentation. We investigated both full supervision (Fig.~\ref{fig_full_sv}) and mixed supervision, where one fully segmented image was used in addition to the prior rules (Fig.~\ref{fig_mixed_sv}). 
Under full supervision with a set of groundtruth edge weights, we compute the Dice Score \cite{DBLP:journals/corr/abs-1906-11031} of the predicted edge weights $a$ and the ground-truth $\hat{a}$ for each sub-graph and use it as the reward. We find this approach to be robust against class imbalance.
In both cases, the agent learns to segment the circles correctly, demonstrating fast and robust convergence. Note that learned pixel features converge to a state which strongly resembles a semantic segmentation of the image.
We have also used the mixed-supervision approach in Subsection \ref{sec_exp_bio}.

\subsection{Synthetic Circle Experiments} \label{sec:circle_rewards}

For the synthetic circular data we implement a reward function based on the circular hough transform (CHT).
The object rewards $r_{fg}$ are computed as follows: we define a threshold $\gamma$ for the CHT value (Fig.~\ref{fig_reward_landscape} shows the reward surface for $\gamma=0.8$). 
Let $c\in [0, 1]$ be the CHT value for to the given object, let $k$ be the number of expected objects and $n$ be the number of predicted objects. We then define the local and global reward per object as follows.

\begin{align} \label{eq:reward_circles}
    r_{locacl}&= 
    \begin{cases}
        \sigma \left( (\frac{c - \gamma}{1 - \gamma} - 0.5) \, 6 \right) \, 0.4,& \text{if } c\geq \gamma\\
        0,              & \text{otw}
    \end{cases}\\
    r_{global}&= 
    \begin{cases}
        0.6 \, \left( \frac{k}{n} \right),& \text{if } n\geq k\\
        0.6,              & \text{otw}
    \end{cases}\\
    r_{fg} &= r_{local} + r_{global}
\end{align}
Here $\sigma(\cdot)$ is the sigmoid function. The input to $\sigma(\cdot)$ is normalized to the interval $[-3, 3]$. The rewards are always in $[0, 1]$, the local reward is in range $[0, 0.4]$ and the global reward is in range $[0, 0.6]$.

We assume that the largest object is the background and assign it the background reward:
\begin{align} 
    r_{bg} &=
    \begin{cases}
        \left( \frac{n}{k} \right),& \text{if } n\le k\\
        1,              & \text{otw}
    \end{cases}
\end{align}

\subsection{DSB Experiments} \label{sec:dsb}

We compute the per-object rewards based on the following properties: the eccentricity, the extent, the minor axis length, the perimeter, the solidity as well as the mean, maximal and minimal intensity. The properties are computed using the "regionprops" functionality from \href{https://scikit-image.org/}{skimage}. Using these properties $p_i$ and their expected values $\hat{p}_i$ determined for simplicity from ground-truth objects, we compute the reward $r_o$ as
\begin{equation}
    r_o =\frac{1 + cs(\frac{p_i}{n_i}, \frac{\hat{p}_i}{\hat{n}_i})}{2}.
\end{equation} 
Here, $n_i$ and $\hat{n}_i$ are normalization factors to bring each property into the range $[0,1]$ and $cs(\cdot, \cdot)$ is the cosine similarity.
Note that objects that are identified as background, using a simple size criterion, do not receive any reward.
Projecting this reward to edges using the maximum (see Section \ref{sec:subgraph_supervision}) yields $r_{oe}$, and the final edge-level reward $r_e$ is constructed by the sum 
\begin{equation}
    r_e =\alpha \, r_{oe} + \beta \, \hat{r}_{e}.
\end{equation} 
Here $\alpha$ and $\beta$ are scaling factors and $\hat{r}_{e}$ is a reward based on the distance of the action to the difference of the mean intensity $\mu$ of the incident superpixels of edge $(i, j)$
\begin{equation}
    \hat{r}_{e, ij} = |(1-a_{ij}) - (|\mu_{i} - \mu_{j}|)|.
\end{equation} 
If the action for an edge is close to $1$, i.e. strongly favoring a split, and the difference in the mean intensity is high, there will be a large reward and vice versa.

For the DSB experiments we use three sets of superpixels:
\begin{itemize}
    \item ``GT'': these superpixels take into account ground-truth information. They are used to judge the performance of our method without being limited by the quality of the underlying superpixels. For these superpixels we compute a height map based on the gradient of the input image and then perform a seeded watershed transform with seeds at the minima of the height map. The resulting superpixels are intersected with the groundtruth, further breaking into pieces the superpixels which cover more than one object or that spill into the background.
    \item ``UNET'': these superpixels are based on predictions from a pre-trained U-Net and allow the performance of our method when having access to such a pre-trained network. The U-Net predicts foreground and boundary probabilties and the superpixels are computed via a watershed that uses the minima of the boundary probabilities as seeds and also uses these probabilties as heightmap. Befor computing the seeds the boundary probability are smoothed, using a higher smoothing factor in the background than in the foreground, which is determined based on the foreground predictions.
    \item ``RAW'': these superpixels are computed based on raw image data only. They are computed similar to the ``UNET'' superpixels, but the gradient image of the input is used to compute seeds and as heightmap instead of the boundary predictions. To determine foreground vs. background the otsu thresholded intensity image is used.
\end{itemize}
Furthermore, we use groundtruth objects to determine expected values of the priors. In practice, these are usually known from biology (shape priors) or can be determined from bulk measurements without segmentation (intensity priors).

\begin{figure*}
	\centering
	\includegraphics[width=1.\textwidth]{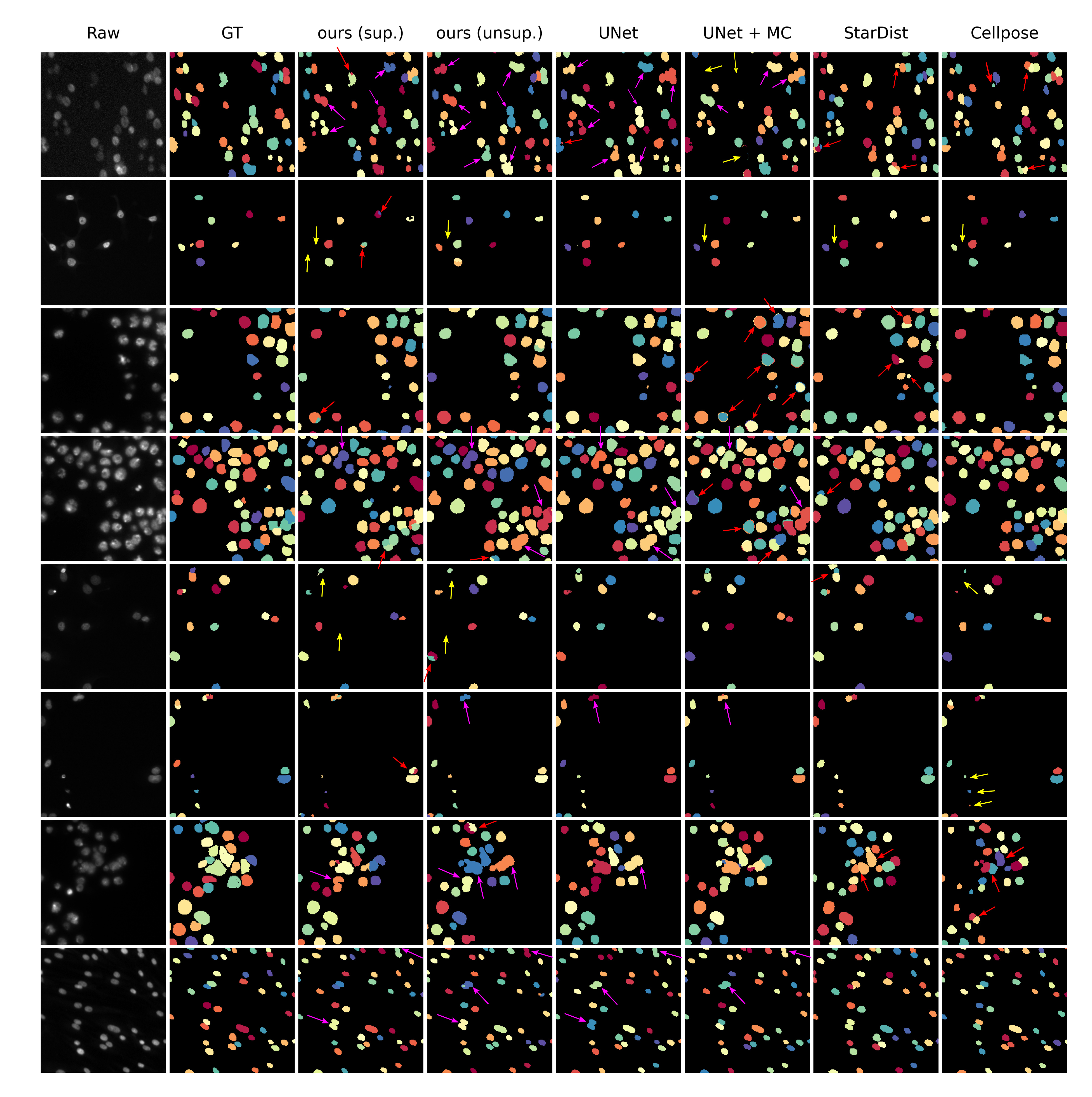}
	\caption{Example segmentation results for our method and baselines. The arrows mark segmentation errors: red are false splits, purple are false merges and yellow are omissions or nuclei segmented with a very wrong shape. Note that the errors were annotated manually to give an impression of the different kind of systematic errors and are not exhaustive for these images.}
	\label{fig_dsb_results}
\end{figure*}

Fig.~\ref{fig_dsb_results} shows segmentation results for the methods from Tab.~\ref{tab:dsb_results} for several images from the test set.
Here, red arrows mark segmentation errors that wrongly split a nucleus, purple arrows mark segmentation errors that wrongly merge nuclei and yellow arrows mark segmentation errors that either omit nuclei or segment them
with a completely wrong shape. Note that these errors were manually annotated and are not exhaustive for the shown images.
We observe that different methods suffer from different systematic errors:
Our proposed method suffers from merges, sometimes omits nuclei and in some cases wrongly splits of a small superpixel from a nucleus.
The UNet predominantly suffers from merges. The combination of UNet and Multicut, which uses the same superpixels as our method, suffers from merges and omissions, it also systematically splits off superpixels located on the boundary of nuclei, which can especially be seen for the images in the third and fourth row.
Stardist and Cellpose, which use use a strong shape prior, do not suffer from merges. Instead, they often wrongly split up nuclei that do not adhere to the shape prior and sometimes omit nuclei or predict them with a very wrong shape. Note that both methods with shape priors result in round shapes that may not match the ground-truth objects for high intersection thresholds, even if they are visually matching well.

